\documentclass{article}
\usepackage{spconf,amsmath,graphicx}

\usepackage{algorithmic}
\usepackage{algorithm}
\usepackage{amsmath}
\usepackage{amsfonts}
\usepackage{hyperref}

\title{An Efficient DP-SGD Mechanism for Large Scale NLU Models}
\name{$\text{Christophe Dupuy}^{1}$, $\text{Radhika Arava}^{1}$, $\text{Rahul Gupta}^{1}$, $\text{Anna Rumshisky}^{1, 2}$}
\address{$\:^{1}$Amazon Alexa AI, Cambridge, MA, USA\\ 
$\:^{2}$University of Massachusetts, Lowell, MA, USA}
\begin{document}
%
\maketitle
\begin{abstract}
Recent advances in deep learning have drastically improved performance on many Natural Language Understanding (NLU) tasks. However, the data used to train NLU models may contain private information such as addresses or phone numbers, particularly when drawn from human subjects.
It is desirable that underlying models do not expose private information contained in the training data. 
Differentially Private Stochastic Gradient Descent (DP-SGD) has been proposed as a mechanism to build privacy-preserving models.
However, DP-SGD can be prohibitively slow to train.
In this work, we propose a more efficient DP-SGD for training using a GPU infrastructure and apply it to fine-tuning models based on LSTM and transformer architectures. 
We report faster training times, alongside accuracy, theoretical privacy guarantees and success of Membership inference attacks for our models and observe that fine-tuning with proposed variant of DP-SGD can yield competitive models without significant degradation in training time and improvement in privacy protection.
We also make observations such as looser theoretical $\epsilon, \delta$ can translate into significant practical privacy gains.
\end{abstract}
\section{Introduction}

Large scale NLP models have contributed significantly to the success of commercial voice assistants like Amazon Alexa, Google Assistant and Siri. 
They have shown high generalization accuracies for various learning tasks, from question answering \cite{qu2019bert} to named entity recognition \cite{akbik2019pooled}.
However, large NLP models can be prone to privacy attacks (e.g. MIA: Membership Inference Attacks \cite{shokri2017membership}) and can leak data used to train these models. 
In this paper, we focus on measuring and mitigating the privacy risks of these models. 
Specifically, differentially private (DP) model building algorithms have shown promise in providing defense against privacy attacks \cite{erlingsson2019amplification, rahimian2019differential}. 
We focus on a specific central differential private mechanism - Differentially Private Stochastic Gradient Descent (DP-SGD) and evaluate its impact on model utility and privacy. 

DP-SGD \cite{DPSGD} is an extension over the popular stochastic gradient descent algorithm that offers theoretical ($\epsilon, \delta$) privacy guarantees \cite{Dwork2006}.
In our work, we make extension to proposals by \cite{McMahan2016FederatedL} and \cite{Yu0PGT19} and, propose {\it efficient DP-SGD} (eDP-SGD) suited for GPU training. Specifically, we utilize the noise addition on gradients over data batches proposed by \cite{McMahan2016FederatedL} to compute gradients per GPU for enhanced efficiency and clip parameters for each layer of the NLU models used in this study. Inspired from \cite{Yu0PGT19}, we apply noise decay on gradients computed on each GPU. 
We apply the combination of these techniques to NLU model fine-tuning (as opposed to full training) on dataset of various sizes.
For privacy evaluation, we study the impact of our extension of DP-SGD on MIA success. 
Previous work on the effects of DP on MIA performance study the two extreme cases where either 1) the DP-model is resilient to MIA but suffers significant degradation in performance \cite{rahman2018membership}; or 2) the DP-model achieves similar performance to the non-private model but does not offer significant gains in privacy \cite{jayaraman2019evaluating}. We observe that applying our method with looser theoretical DP guarantees translate into significant reduction in MIA performance.
We also report the impact of using DP-SGD and our extension on the training time which, to the best of our knowledge, no prior work has done. We observe that DP-SGD can be prohibitively slower (up to 150 times) than non-private baselines while our method, in the worst case, is slower by a factor of 2.

In this paper, we make the following contributions: (i) We study the impact of applying DP techniques for NLP models trained on large-scale datasets. We present a computationally efficient setting for DP-SGD and provide a comparison in terms of training time, accuracy and privacy of the non-private and DP models. No other study has done this comparison, let alone for large-scale NLP setting. (ii) We demonstrate that using DP-SGD during fine-tuning, one can obtain models with competitive utility (in comparison to models trained with vanilla SGD), while achieving significant gains in protection against privacy attacks and, (iii) Building on existing DP-SGD variants, we propose an extended version of the DP-SGD technique which is computationally-efficient and report significant compute gains over DP-SGD.  

\section{Related work}
\label{sec:bckgd}

DP-SGD \cite{DPSGD} modifies vanilla SGD by clipping the gradients computed over each individual datapoint, followed by accumulation of the clipped gradients over a batch and noise addition.
Researchers have applied DP-SGD to reduce memorization in language models \cite{mcmahan2017learning}, data generation \cite{xie2018differentially} and image classification \cite{rahman2018membership}.
\cite{chen2020understanding} aim to understand the properties of DP-SGD.
Attempts have been made to improve efficiency of DP-SGD by improving communication protocols in a distributed training setting, where individual servers contribute gradient on locally stored data and privacy of data in each local server is desirable 
\cite{agarwal2018cpsgd}.
Adaptive variants of DP-SGD in a federated setting have also been proposed \cite{thakkar2019differentially}.

MIA has been studied in a variety of settings, such as MIA with synthetic or noisy data \cite{shokri2017membership}, shallow models \cite{mia2019MLAAS}, non-matching shadow and target data sets \cite{Salem2019MLLeaksMA}. 
DP-SGD while carrying theoretical privacy guarantees has also been demonstrated to provide defense against MIA \cite{rahman2018membership}.
In this work, we report both theoretical and MIA based privacy quantifications.

\section{Efficient DP-SGD}
\label{sec:method}
In this section, we present the eDP-SGD algorithm by modifying the following three techniques and, adapting them for a GPU based training. Given the high risks associated with NLP models revealing private information, and the growing concern over these  risks, as well as the scarcity of studies in the domain of privacy-preserving algorithms for NLP, we have chosen some sensitive and proprietary datasets of a voice assistant for our studies. We propose the following modifications to the existing DP-SGD technique.

{\bf Micro-batch computations per GPU}
\label{sec:microbatch}
DP-SGD requires clipping the gradient of every single example in the batch.
This induces a significant computational cost since given a batch size $B$, DP-SGD would require the computation of $B$ gradients, as opposed to computation of one gradient in classic SGD. 
\cite{McMahan2018AGA} show that it is possible to group examples in \textit{micro-batches} in the DP-SGD scheme and still maintain DP-guarantees for the resulting model. 
This manipulation is equivalent to a global Gaussian mechanism and the authors provide the equivalence relationship in their paper.
We leverage this work and apply DP-SGD computations to the micro-batch contained within each GPU. 
Given a batch of data points and $n$ GPUs, we divide the batch into $n$ micro-batches, processed independently on each GPU. 

We make another addition to the algorithm suggested by \cite{McMahan2018AGA}, and add scaled noise to gradient computed per micro-batch within the GPU (as opposed to adding noise post the aggregation of gradients from all micro-batches). 
Given $n$ GPUs, adding a Gaussian noise $\mathcal{N}(0, \sigma^2)$ to gradients per micro-batch is equivalent to adding a noise $\mathcal{N}(0, (n\sigma)^2)$ to the gradients aggregated over the micro-batch.
This change relaxes the need for aggregation before noise addition and accelerates computation.

{\bf Scaling}
\label{sec:scaling}
For large scale models, the magnitude of the gradient varies across parameters in the model. 
For instance, the magnitude of gradients for parameters in the lower layers can be different compared to those in the upper layers.
Hence, using a constant clipping parameter $C$ can either be too aggressive or too weak for a certain set of parameters. 
In this case, given a set of parameters $w^k$ (e.g. those drawn from the $k^\text{th}$ layer in a neural network), a strategy that clips gradients with a parameter ($C_k$) specific to the set $w^k$ is preferable.
However, this strategy may again lead to poor privacy guarantees for large variations in $C_k$ assigned to each set of parameters. 
Inspired by the scaling approach suggested by \cite{McMahan2018AGA}, we compute a scaling factor for each layer, proportional to the norm of gradient calculated on the first iteration for each layer. 
The scaling is applied to gradient computed over a micro-batch contained in each GPU.
Since the model parameters are randomly initialized and that the gradient norm is expected to decrease during training, the norm of the gradient in the first iteration gives a rough estimation of the upper bound of the gradient magnitude throughout training. 
We scale a constant clipping value $C$ by the factor $\alpha_k$ for each layer.
This strategy also reduces the number of hyper-parameters to tune for every new model or dataset. 

{\bf Noise Decay}
\label{sec:decay}
In DP-SGD, the amount of noise added is the same for all the training iterations with a variance equal to the clipping parameter $C$ times the noise multiplier $z$, used  to compute theoretical DP guarantees. 
As the magnitude of the gradients is likely to decrease when approaching convergence, adding noise with constant variance can lead to slower convergence \cite{Yu0PGT19}; as the magnitude of the noise can be significantly higher than the magnitude of the gradient. 
In addition, a noise with high variance can wash out the information contained in the gradients after a few epochs, while a noise with low variance would yield low privacy guarantees. 
To improve convergence, \cite{Yu0PGT19} use noise variance reduction at every epoch by scaling the initial noise multiplier $z_0$ by a decreasing function $d_\tau$ (parameterized by $\tau$). 
In our work, we use this strategy, however, it is applied to gradients computed independently at each GPU.
We decay the noise strength as a function of epoch number $t$ using one of the following forms of the multiplier $d_\tau$: 
\begin{enumerate}
    \item Linear decay: $d_\tau = 1/(1 + \tau t)$ 
    \item Exponential decay: $d_\tau = e^{-\tau t}$
\end{enumerate}

Algorithm \ref{algo:dpsgd} summarizes eDP-SGD.

\begin{algorithm}[t]
\caption{eDP-SGD}
\label{algo:dpsgd}
\begin{algorithmic}
\REQUIRE {\textbf{GPU Devices:} $\mathrm{GPU}_1,\ldots, \mathrm{GPU}_{N}$; \\ \textbf{Data:} Batch $B=(x_1,\ldots,x_{\vert B\vert})$;~$\vert B\vert > N$. \\ \textbf{DP-SGD Input:} Noise multiplier~$z_0$; Clipping coefficient~$C$; Noise decay $d:\mathbb{N}\mapsto\mathbb{R}_+^*$; Scaling $(\alpha_k)_k$;\\ \textbf{Model Input:} Loss $L(\cdot, w)$; Epoch $t$;}
\ENSURE DP-gradient for optimizer
\FOR{$(M, \mathrm{GPU})$ in $\left\{(M_i,\mathrm{GPU}_i)\right\}_{i=1,\ldots,N}$}
\STATE  Send micro-batch $M$ to $\mathrm{GPU}$
\STATE Compute gradient:\\ \;\;\;$\Delta^M \gets \nabla_w\left(\frac{1}{|M|}\sum_{x \in M} L(x,w)\right)$
\STATE Scale: $\forall k,\, \Delta^M_k \gets \Delta^M_k / \alpha_k $
\STATE Clip: $\Delta^M \gets \min\left(1, \frac{C}{\Vert\Delta^M\Vert_2}\right)\Delta^M$
\STATE Rescale: $\forall k,\,\Delta^M_k \gets \alpha_k \Delta^M_k$
\STATE Set noise multiplier: $z_t \gets z_0\cdot d(t)$
\STATE Add scaled Gaussian noise: \\ \;\;\;$\Delta^M \gets \Delta^M + \frac{y}{N}$, $y\sim\mathcal{N}(0,(C\cdot z_t)^2)$ 
\ENDFOR
\STATE Aggregate: $\Delta\gets \frac{1}{N}\sum_M \Delta^M$ 
\STATE Return $\Delta$
\end{algorithmic}
\end{algorithm}

\section{Experimental setup}
\label{sec:expes}
We focus on Intent Classification (IC) and Named-Entity Recognition (NER) tasks in this work as they are popular in industrial NLP systems \cite{su2018re}. We next describe the datasets used in our experiments.

\subsection{Datasets} 
We use three publicly available datasets - ATIS \cite{atis1990}, SNIPS \cite{snips2018} and NLU-EVAL \cite{nlueval2019} and three additional datasets from a leading smart-home company: Communication, Health and Video, each containing roughly 1 million utterances. 
Example utterances in the internal datasets are: Communication ``call my parents at 0123'', Health: ``refill my aspirin prescription'' and Video: ``play my favorite movie''. 

We construct a train/validation/test split for these datasets so that the split ratio is approximately the same (45:5:50).
A roughly equal number of datapoints in the train and test set helps us create a balanced evaluation set for training the MIA models. 
The ``member'' utterances used to train/evaluate the MIA success are sourced from the IC-NER training sets, and an equal number of ``non-member'' utterances are sourced from the test set (not used in training). 

 \begin{table*}[t]
     \centering
     	\caption{Results showcasing privacy-utility tradeoff of NLU models trained using eDP-SGD against models trained with SGD. Datasets are arranged by size. $\Delta$SER and $\Delta$MIA represent relative changes w.r.t baseline. $\delta$ is set to $5 \times 10^{-4}$ for public corpora and $5 \times 10^{-5}$ for Alexa datasets during computation of $\epsilon$ using $z$CDP \cite{Yu0PGT19}}
     \label{tab:results_acc}
     \begin{tabular}{l|rr|rrr|rrr|rrr}
         \hline
          \bf Dataset & \multicolumn{2}{|c|}{\bf SGD} & \multicolumn{3}{|c|}{\bf No decay} & \multicolumn{3}{|c|}{\bf Linear Decay} &  \multicolumn{3}{|c}{\bf Exponential Decay} \\ \hline 
          & SER & MIA-AUC & $\Delta$SER & $\Delta$MIA & $\log_{10}\epsilon$ &  $\Delta$SER & $\Delta$MIA & $\log_{10}\epsilon$ & $\Delta$SER & $\Delta$MIA & $\log_{10}\epsilon$  \\ \hline
          	 \multicolumn{12}{c}{\bf CLC model}\\ \hline
          ATIS  & 6.3 & 67.5 & -12.9 & -8.26 & 4.4 & -10.6 & -11.6 & 5.8 & -11.9 & -11.3 & 24.4 \\ 
          SNIPS   & 13.0 & 70.6 & -1.8 & -2.5 & 3.1 & -10.3 & 1.8 & 6.5 & -6.7 & 4.4 & 17.4 \\ 
          NLU-eval  & 26.7 & 65.3 & -0.7 & -1.4 & 2.0 & 3.6 & -7.6 & 2.7 & 1.4 & -4.6 & 22.1 \\ 
        Health  & - & - & -1.2 & -3.1 & 3.6 & -2.5 & -2.3 & 4.0 & -2.7 & -1.5 & 4.3 \\ 
          Comm.  & - & - & 0.3 & -3.2 & 6.2 & 0.3 & -5.7 & 7.5 & 0.5 & -5.4 & 19.1 \\ 
          Video & - & - & 4.0 & -2.4 & 4.8 & 4.4 & -3.7 & 5.9 & 3.1 & -6.0 & 19.5 \\ 
	\hline
	 \multicolumn{12}{c}{\bf BERT model}\\ \hline
          ATIS   & 5.4 & 56.1 & 1.3 & -2.5 & 5.9 & -9.7 & -5.2 & 6.3 & -4.1 & -5.8 & 11.1 \\ 
          SNIPS & 12.7 & 65.3 & -21.5 & -8.3 & 2.7 & -20.3 & -9.2 & 2.3 & -7.1 & -8.3 & 2.7 \\ 
          NLU-eval & 24.1 & 63.6 & -2.4 & -7.6 & 2.2 & 4.6 & -11.6 & 1.7 & 0.3 & -9.5 & 2.3 \\ 
          Health   & - & - & 3.2 & -4.1 & 1.9 & 0.8 & -2.9 & 2.8 & 1.2 & -4.3 & 2.9 \\ 
          Comm. & - & - & 0.3 & -4.4 & 2.1 & 0.2 & -3.4 & 2.8 & 0.7 & -4.4 & 3.7 \\ 
          Video  & - & - & 4.7 & -10.1 & 1.9 & 4.6 & -10.9 & 1.8 & 3.3 & -8.2 & 2.3 \\ 
	\hline

     \end{tabular}
     
 \end{table*}

\begin{table}[t]
     \centering
    \caption{Comparison of training time per epoch for SGD, DP-SGD and eDP-SGD.}
     \label{tab:results_time}
     \begin{tabular}{@{}l|c|c|c@{}}
         \hline
               \bf Dataset & {\bf SGD} & {\bf DP-SGD} &  \bf eDP-SGD \\  \hline
    &  Time/epoch & \multicolumn{2}{c}{Multiplicative factor}\\ \hline
	\multicolumn{4}{c}{\bf CLC model}\\ \hline
          ATIS & 9.5s & $\times$2.1 & $\times$1.04 \\ 
          SNIPS & 9.8s & $\times$5.1 & $\times$1.01 \\ 
          NLU-EVAL & 9.9s & $\times$ 7.3 & $\times$1.03\\
                    Health & 21.8s & $\times$153.9 & $\times$1.14 \\
          Comm. & 109.8s & $\times$143.5 & $\times$1.15 \\
          Video & 83.7s & $\times$152.4 & $\times$1.2 \\
          \hline
    \multicolumn{4}{c}{\bf BERT model}\\ \hline
          ATIS & 2.5s & $\times$13.0 & $\times$1.29  \\ 
          SNIPS & 2.8s & $\times$22.8 & $\times$1.52 \\ 
          NLU-EVAL & 3.5s & $\times$29.2 & $\times$1.62  \\ 
                    Health & 56.5s & $\times$82.0 & $\times$2.06 \\
          Comm. & 398.8s & $\times$73.8 & $\times$2.05 \\
          Video & 276.0s & $\times$80.3 & $\times$2.17 \\
	\hline
	\end{tabular}
	 
 \end{table}

\subsection{Models} 
We train IC and NER models using the following two architectures. These architectures are chosen to capture training on an LSTM based and a transformer based model. These architectures have consistently pushed state of the art on several tasks \cite{devlin2018bert, peters2018deep}. 

{\bf CLC} \cite{CLC2016}: The CLC model takes as input the concatenation of a token-level character CNN output with token embeddings, followed by a bi-LSTM layer. IC and NER models are a fully connected layer and a CRF layer, respectively, on top of the bi-LSTM layer.
We use pre-trained FastText embeddings \cite{mikolov2018fasttext} for inputs of our models. In order to train IC-NER models on the public data, we use the pre-trained embeddings obtained using Wiki-news corpus \footnote{\href{https://fasttext.cc/docs/en/english-vectors.html}{https://fasttext.cc/docs/en/english-vectors.html}}. 
In order to train the models on the internal corpora, we train FastText embeddings on a separate set of utterances sourced from the same smart home agent.
We use two bi-LSTM layers, each with a hidden dimension of 384.

{\bf BERT}: \cite{devlin2018bert}
We also train IC-NER layers on top a BERT model, pretrained on a combination of Wikipedia articles dataset \footnote{\href{https://www.wikipedia.org}{https://www.wikipedia.org}} and One Billion Word corpus \cite{chelba2013onebillion} using the masked language modeling task.
Our BERT model has 4 layers, 12 attention heads, and a hidden dimensions of 312. We use the sum of the CRF loss (NER) and the cross entropy loss (IC) as the optimization objective.

We tuned the value of $C$ from the set $\{0.1, 0.25, 0.5, 1, 2, 5\}$, $z_0$ from values ranging from $10^{-6}$ to $1$. We tuned $\tau$ from the set $\{0.02, 0.05, 0.07, 0.1, 0.3, 0.5, 0.7, 0.9\}$ for linear decay; and from the set $\{0.01, 0.02, 0.05, 0.08, 0.1, 0.2, 0.35, 0.5\}$ for exponential decay.
We use a p3.16xlarge instance\footnote{\href{https://aws.amazon.com/ec2/instance-types/}{https://aws.amazon.com/ec2/instance-types/}} with 8 GPUs to train each model. We implemented the model block and the DP scheme with MXnet \cite{chen2015mxnet} and we leverage Horovod \cite{sergeev2018horovod} to boost efficiency.

We report the semantic error rate (SER \cite{su2018re}, a normalized measure for IC-NER task) metric as the utility metric (lower is better).
We also report the theoretical DP $(\epsilon, \delta)$ guarantees (for DP models) and the success of an MIA attack captured using the area under ROC curve (AUC) metric. 
Following \cite{shokri2017membership} structure for MIA, we train a shadow model on a chosen public corpora. The attack model is then trained on the outputs of this shadow model and evaluated on the outputs of the model under attack.
Sorted IC and NER output scores are used as features to attack model.
As a reminder, during MIA evaluation, train set for the model under attack are used as {\it member} set, while the test set is used as {\it non-member} set.
Finally, we also report the time take to train the models using SGD/eDP-SGD/DP-SGD. 
ADAM optimizer is use for all -- SGD/eDP-SGD/DP-SGD settings.

\section{Results}
\label{sec:results}
Table \ref{tab:results_acc} presents the utility/privacy metrics obtained using SGD training, in addition to relative changes in those metrics when trained using various decay schemes in eDP-SGD (for the sake or brevity, we do not report DP-SGD results separately as they are similar to the no decay scheme). 
From the results, we sometimes observe an improvement in performance (particularly for public corpora), as was also reported by \cite{khatri2017preventing}. 
This is encouraging and we attribute the improvement in performance to eDP-SGD acting as a regularizer.
For smaller datasets, it is easier to overfit the CLC and BERT models to the training set.
On the larger corpora, the model error rates does not degrade significantly (except for Video).
Additionally, either linear or exponential decay offer a better privacy-utility trade-off than no decay on larger datasets where a loss in utility is expected (we embolden a lower utility loss and higher MIA value compared to no decay for communication and video datasets in Table \ref{tab:results_acc}). 
This indicates that decay when applied independently at each micro-batch still improves privacy-utility trade-off, but the observation is limited to larger datasets.
We also observe that the theoretical $\epsilon$ privacy guarantee seem to have a loose correlation with MIA success rate, where sometimes higher value of $\epsilon$ is associated with greater decrease in MIA success rate.
Table \ref{tab:results_time} also demonstrates that our algorithm does not degrade the training time over SGD. On the other hand, DP-SGD can increase training time by factor of up to 150.
Overall, these results indicate that task specific fine tuning with eDP-SGD can be used in industrial settings.

\section{Conclusion}
\label{sec:ccl}
In this work, we propose a variant of DP-SGD that is suited for GPU based training and use it during fine-tuning IC-NER models on multiple datasets.
We report training time, accuracy and privacy metrics on two model architectures, and argue that task specific fine-tuning with eDP-SGD is practical for large scale model training from an accuracy and efficiency perspective.
We make observations such as our methods can provide a better utility privacy trade-off and provides significant gains in the training time. 
We also observe that practical MIA guarantees see improvement even when the theoretical $\epsilon$ values are high in value. 

In the future, one can explore the protection offered by DP-SGD against privacy attacks on recently proposed larger models.
We also aim to further analyze the weak correlation between theoretical guarantees and MIA.
We can also study the impact of other techniques like regularization and quantization (alongside DP-SGD) on model memorization.

\bibliographystyle{IEEEbib}
\bibliography{short_bib}

\end{document}